%% file: main_v1.tex
\documentclass[10pt,twocolumn,letterpaper]{article}

\usepackage[pagenumbers]{cvpr}

\usepackage[accsupp]{axessibility}
\input{preamble}

\definecolor{cvprblue}{rgb}{0.21,0.49,0.74}
\usepackage[pagebackref,breaklinks,colorlinks,allcolors=cvprblue]{hyperref}

\newcommand{\blfootnote}[1]{%
  \begingroup
  \renewcommand{\thefootnote}{}%
  \footnote{#1}%
  \addtocounter{footnote}{-1}%
  \endgroup
}

\def\confName{CVPR}
\def\confYear{2026}

\title{Experience Transfer for Multimodal LLM Agents in Minecraft Game}

\author{
Chenghao Li$^{1}$,
Jun Liu$^{1}$,
Songbo Zhang$^{1}$,
Huadong Jian$^{1}$,
Hao Ni$^{2}$,\\
Lik-Hang Lee$^{3}$,
Sung-Ho Bae$^{4}$,
Guoqing Wang$^{1}$,
Yang Yang$^{1}$,
Chaoning Zhang$^{1*}$
}

\begin{document}

\twocolumn[{%
\renewcommand\twocolumn[1][]{#1}%
\maketitle
\vspace{-3em}
\begin{center}
    \captionsetup{type=figure}
    \includegraphics[width=\linewidth]{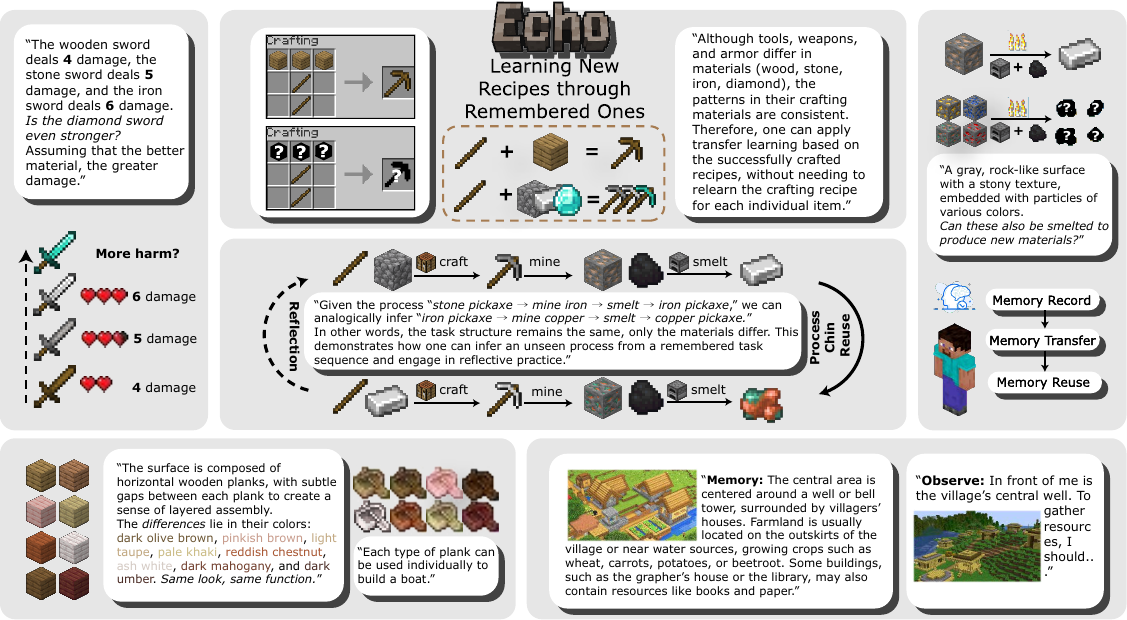}
    \captionof{figure}{\textbf{Conceptual illustration of Echo.}
    The agent learns from experience and discovers transferable patterns, enabling interpretable analogy-based reasoning and cross-task generalization. In some classical methods, such as DEPS~\citep{wang2023describe} and JARVIS-1~\citep{wang2024jarvis}, ICL is mainly used to retrieve few-shots from the memory bank to assist in generating sub-task sequences for the current goal. Echo, on the other hand, treats ICL learning as an active process — it proactively retrieves potentially new tasks from the memory bank for validation and execution.}\label{fig:concept}
\end{center}%
}]

\blfootnote{$^{*}$ Corresponding author. $^{1}$University of Electronic Science and Technology of China;
$^{2}$KAIST;
$^{3}$The Hong Kong Polytechnic University;
$^{4}$Kyung Hee University.
Project: https://github.com/CatworldLee/Echo
}

\input{sec/0_abstract}
\input{sec/1_intro}
\input{sec/2_relat}
\input{sec/3_method}
\input{sec/4_experiment}
\input{sec/5_conclusion}

{
    \small
    \bibliographystyle{ieeenat_fullname}
    \bibliography{main}
}

\end{document}

%% file: preamble.tex
\usepackage{multirow, multicol}
\usepackage{array}
\usepackage{booktabs}
\usepackage{colortbl}
\usepackage{xcolor}
\usepackage{fontawesome5}

%% file: sec/0_abstract.tex
\begin{abstract}
Multimodal LLM agents operating in complex game environments must continually reuse past experience to solve new tasks efficiently. In this work, we propose Echo, a transfer-oriented memory framework that enables agents to derive actionable knowledge from prior interactions rather than treating memory as a passive repository of static records. To make transfer explicit, Echo decomposes reusable knowledge into five dimensions: structure, attribute, process, function, and interaction. This formulation allows the agent to identify recurring patterns shared across different tasks and infer what prior experience remains applicable in new situations. Building on this formulation, Echo leverages In-Context Analogy Learning (ICAL) to retrieve relevant experiences and adapt them to unseen tasks through contextual examples. Experiments in Minecraft show that, under a from-scratch learning setting, Echo achieves a 1.3$\times$--1.7$\times$ speed-up on object-unlocking tasks. Moreover, Echo exhibits a burst-like chain-unlocking phenomenon, rapidly unlocking multiple similar items within a short time interval after acquiring transferable experience. These results suggest that experience transfer is a promising direction for improving the efficiency and adaptability of multimodal LLM agents in complex interactive environments.
\end{abstract}

%% file: sec/1_intro.tex
\section{Introduction}
\label{sec:intro}

Driven by embodied intelligence and complex interactive tasks, planning-oriented multimodal large language model (MLLM) agents have rapidly emerged~\citep{wang2023describe,wang2023voyager,park2025mrsteve,qin2024mp5,wang2024jarvis,zheng2024steveeye,zhang2026glosasum}. Representative works such as Voyager~\citep{wang2023voyager} and JARVIS-1~\citep{wang2024jarvis} in environments like Minecraft demonstrate open-ended exploration driven by a ``perception-reasoning-action-memory'' loop, where agents continuously self-improve through environmental feedback. These systems reveal the potential to decompose goals, plan subtasks, and invoke tools without large-scale task-specific supervision. Such agents~\citep{wang2023voyager,wang2024jarvis} typically integrate chain-of-thought reasoning from language models with environment interaction trajectories to generate executable strategies, codes, and action sequences---paving the way for agents that move from ``speaking'' to ``doing.''

Alongside this trend, the demand for task memory and reasoning-based planning becomes particularly prominent. Long-term persistent memory enables an agent to reuse skills and support reasoning about the current task in complex scenarios~\citep{wang2023voyager,wang2024jarvis,park2023generative}. Cross-modal perception enables fine-grained scene understanding~\citep{messina2021fine}. Moreover, explainable reasoning–action coupling allows for hierarchical planning~\citep{ajay2023compositional} and self-verification~\citep{yao2022react} under uncertainty. Existing research has shown that explicit memory structures---such as spatiotemporal event indices~\citep{park2025mrsteve}, multimodal knowledge graphs~\citep{leung2025knowledge}, and structured graph memories~\citep{fu2025vistawise,luo2025gate}---together with retrieval augmentation (extracting examples from historical trajectories, rules, and recipes) can significantly enhance retrieval efficiency, environmental understanding, and long-horizon planning stability, providing support for multi-step reasoning and strategy generation~\citep{wang2023voyager,wang2024jarvis}.

However, despite the promising outlook, most existing methods still treat ``transfer'' only superficially. Memory is often regarded as a passive warehouse~\citep{wang2023voyager,li2024optimus,wang2024jarvis,zhu2023ghost,cai2023groot}, an index of past behaviors, or a library of reusable skills~\citep{wang2023voyager,wang2024jarvis}, while the deeper structures that enable experience to be truly transferable remain underexplored.
Experience transfer lies at the heart of embodied intelligence~\citep{driess2023palm,wang2023voyager}. Once an agent has accumulated sufficient experience and memory, it should be able to infer new knowledge from past experiences---many patterns repeat themselves in subtle, structural ways. Similar landscapes hide similar logic: the same crafting patterns, the same material hierarchies, the same causal chains linking raw resources to useful tools. In worlds like Minecraft, the ``shape prototypes'' of tools and armor, the ``substitutability'' among material families, the common processing chains (gathering → smelting → crafting), and even the functional symmetry between weapons all reveal recurring motifs of transfer. If such transfer axes-shape, material, process, and function---can be explicitly represented and aligned with multimodal embeddings, an agent would no longer need to relearn what it already knows. Instead, it could rapidly adapt, recombine, and repurpose its existing knowledge to thrive in unfamiliar worlds.

To bridge this gap, we introduce a memory-transfer-augmented MLLM planning agent Echo (Figure~\ref{fig:concept}), designed to learn ``how to transfer and when to transfer memory''. The agent decomposes environmental and experiential knowledge into five explicit transfer dimensions---\textit{Structural, Attribute, Procedural, Functional, and Interaction}---forming a unified description of ``what the world is, how it evolves, what can be done, and how to act.''
Within this representation, a Contextual State Descriptor (CSD) encodes visual, textual, and interactive signals into compact, comparable semantic snapshots, aligning multimodal information across the transfer axes.
Finally, a structured in-context analogical learning (ICAL)~\citep{hu2023context} module allows the agent to retrieve, adapt, and build upon past experiences, enabling interpretable reasoning, dynamic self-verification, and knowledge reuse across worlds and tasks.

\begin{figure}[t!]
\centering
\includegraphics[width=\linewidth]{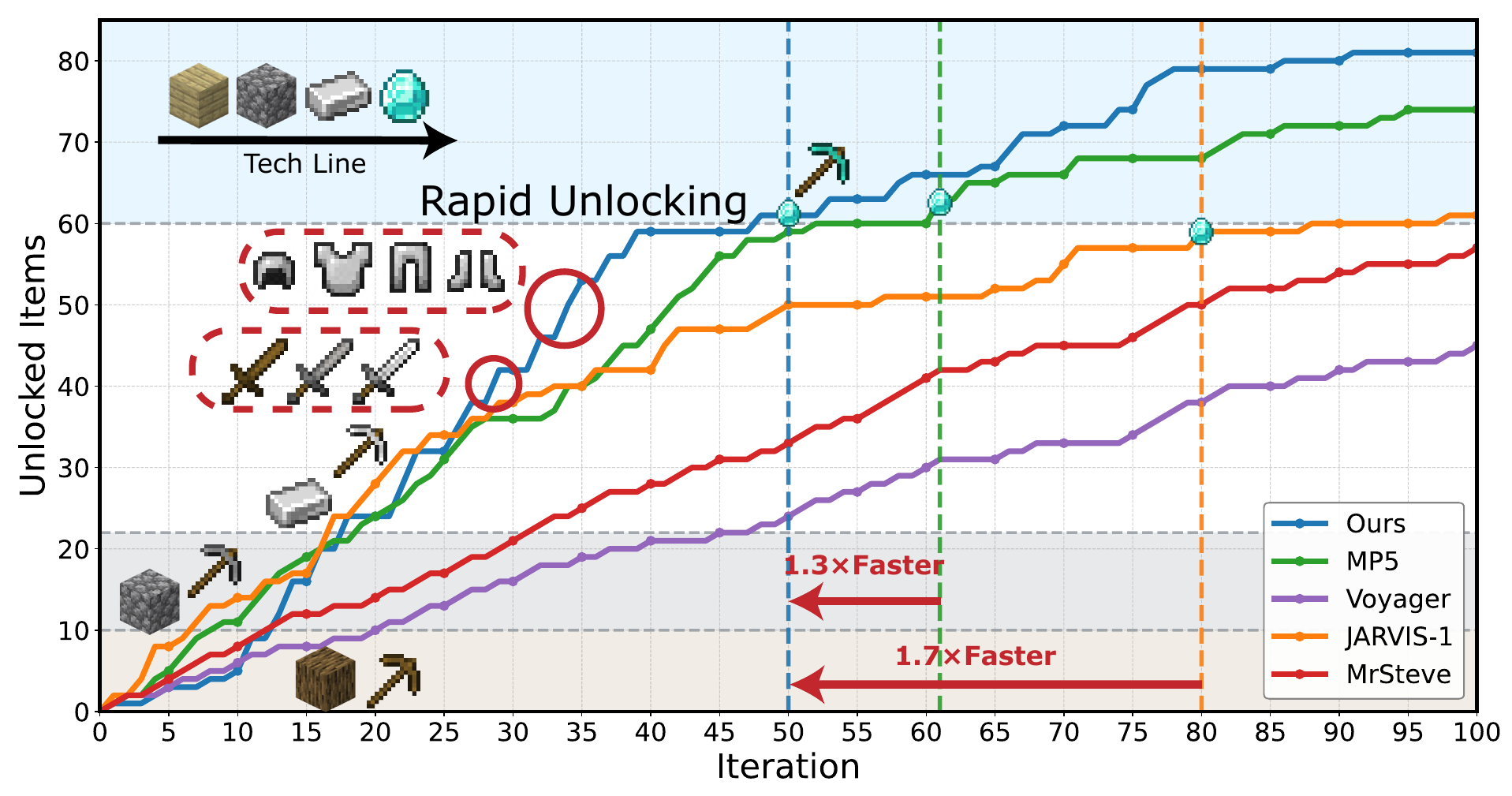}
\caption{\textbf{Comparison of item unlocking progress across different agents.}
The x-axis represents the iteration steps, and the y-axis indicates the number of unique items unlocked.
Our method shows a significantly faster progression, exhibiting a \textit{``rapid unlocking''} phenomenon in the mid-stage, where similar items are unlocked in an explosive manner.
Compared to previous methods (MP5~\citep{qin2024mp5}, Voyager~\citep{wang2023voyager}, JARVIS-1~\citep{wang2024jarvis}, and MrSteve~\citep{park2025mrsteve}), our approach achieves equivalent milestones about \textbf{1.3$\times$--1.7$\times$} faster.}
\label{fig:Unlocking}
\end{figure}

As shown in the Figure~\ref{fig:Unlocking}, after accumulating a certain amount of knowledge during the cold-start phase, our method exhibits a significant increase in learning ability during the mid-to-late phase (item unlocking), resulting in an ``explosive'' item unlocking phenomenon.
In addition, we conducted several experiments to evaluate cross-world adaptation, task generalization, and continuous learning efficiency in complex interactive tasks. The results show that our method, by combining explicit transfer axes, ICAL, self-consistency checking, and memory replay, significantly improves the model's learning efficiency, task success rates, and interpretability, particularly in long-term learning and cross-task transfer scenarios.

Our main contributions are as follows:

\begin{itemize}
    \item To the best of our knowledge, we are the first to explore how to explicitly transfer multimodal memory experiences in multimodal embodied agents.
    \item To address the challenges of structuring and transferring multimodal memory, we propose five explicit transfer dimensions and combine them with ICAL to effectively organize and leverage multimodal memory.
    \item To validate the effectiveness of our approach, we compared it with four agents~\citep{wang2024jarvis,qin2024mp5,wang2023voyager,park2025mrsteve} with different focuses. In learning from scratch tasks, our agent demonstrated clear advantages in learning ability.
\end{itemize}

%% file: sec/2_relat.tex
\section{Related Work}
\label{sec:relat}

\noindent\textbf{Embodied Agents in Minecraft.} Open-world agents learning in complex environments~\citep{guss2019minerl,baker2022video,fan2022minedojo,volum2022craft,wang2023voyager,cai2023open,stengel2024regal,li2024auto,liu2024rl,feng2024llama,zhao2024see,zhao2024hierarchical,dong2024villageragent,zheng2025mcu,yu2025adam,chai2025causalmace,zheng2026llavafa,zhang2026glosasum,zhang2026ghs,zhang2025toward}. MineDojo~\citep{fan2022minedojo} and Voyager~\citep{wang2023voyager} leveraged internet-scale knowledge and large models to achieve open-ended exploration.
Some works enhanced the agents’ active perception capabilities~\citep{zheng2024steveeye,qin2024mp5,zhao2024see}.
The Optimus series~\citep{li2024optimus,li2025optimus} and Jarvis series~\citep{wang2024jarvis,wang2024omnijarvis} adopted modular or hierarchical architectures to enable skill reuse.
GITM~\citep{cai2023groot} and Odyssey~\citep{liu2024odyssey} combined large-scale pretraining with skill libraries to achieve exploratory transfer.
In complex interactive environments (e.g., Minecraft), agents need to possess cross-modal perception and long-term memory capabilities to associate contextual information and reuse past experiences.
For example, MrSteve~\citep{park2025mrsteve} proposed a What-Where-When memory model that supports event retrospection based on temporal–spatial indexing;
VistaWise~\citep{fu2025vistawise} constructed a cross-modal knowledge graph to enable efficient knowledge storage and semantic association;
and the GROOT series~\citep{cai2023groot,cai2024groot} employed graph neural networks to build structured memory.

\noindent\textbf{In-Context Learning.}
In-Context Learning~\citep{dong2024survey,hu2023context} was first systematically introduced in GPT-3~\citep{brown2020language}, referring to the ability to perform new tasks without parameter updates by providing a few demonstrations within the input context. ICL exhibits strong few-shot generalization capabilities.
Existing research mainly focuses on three directions:
(1) \textit{Prompt design and example selection}: improving ICL accuracy and robustness~\citep{liu2022makes,kim2022self,zhou2022large,honovich2023instruction,rubin2022learning,zhang2022active,wang2023large};
(2) \textit{Retrieval-augmented ICL}: combining external knowledge bases to dynamically retrieve demonstrations and mitigate context-length limitations~\citep{chen2024dense,izacard2023atlas,ma2023query,asai2024self};
(3) \textit{Structured ICL}: explicitly modeling structural patterns-e.g., tables, logical chains, or code---to improve transfer~\citep{lu2022fantastically,yang2023not,liu2024context}.
In multimodal domains, models such as CoCa~\citep{yu2022coca} and Flamingo~\citep{alayrac2022flamingo} can rapidly adapt to visual question answering and image understanding with few image–text examples. In interactive environments, ICL facilitates policy and skill transfer; e.g., ReAct~\citep{yao2022react} integrates reasoning traces and action trajectories for context-aware planning.

%% file: sec/3_method.tex
\section{Proposed Method}
\label{sec:method}

\noindent\textbf{Motivation and Challenges.}
In open, interactive, and multimodal environments, our goal is to use retrieval proactively to achieve rapid, stable, and interpretable generalization to new tasks, rather than merely serving current objectives reactively~\citep{wang2023voyager,wang2024jarvis}. At present, we face two major challenges.
First, the real world exhibits highly complex regularities and structural dependencies. Distinct tasks often involve significantly different state transitions and causal relationships, making effective transfer and generalization inherently difficult.
Second, MLLMs tend to suffer from hallucinations and uncontrolled reasoning in open-ended scenarios~\citep{huang2025survey}, undermining both the stability and verifiability of their generalization performance.

\begin{figure}[t!]
    \centering
    \includegraphics[width=\linewidth]{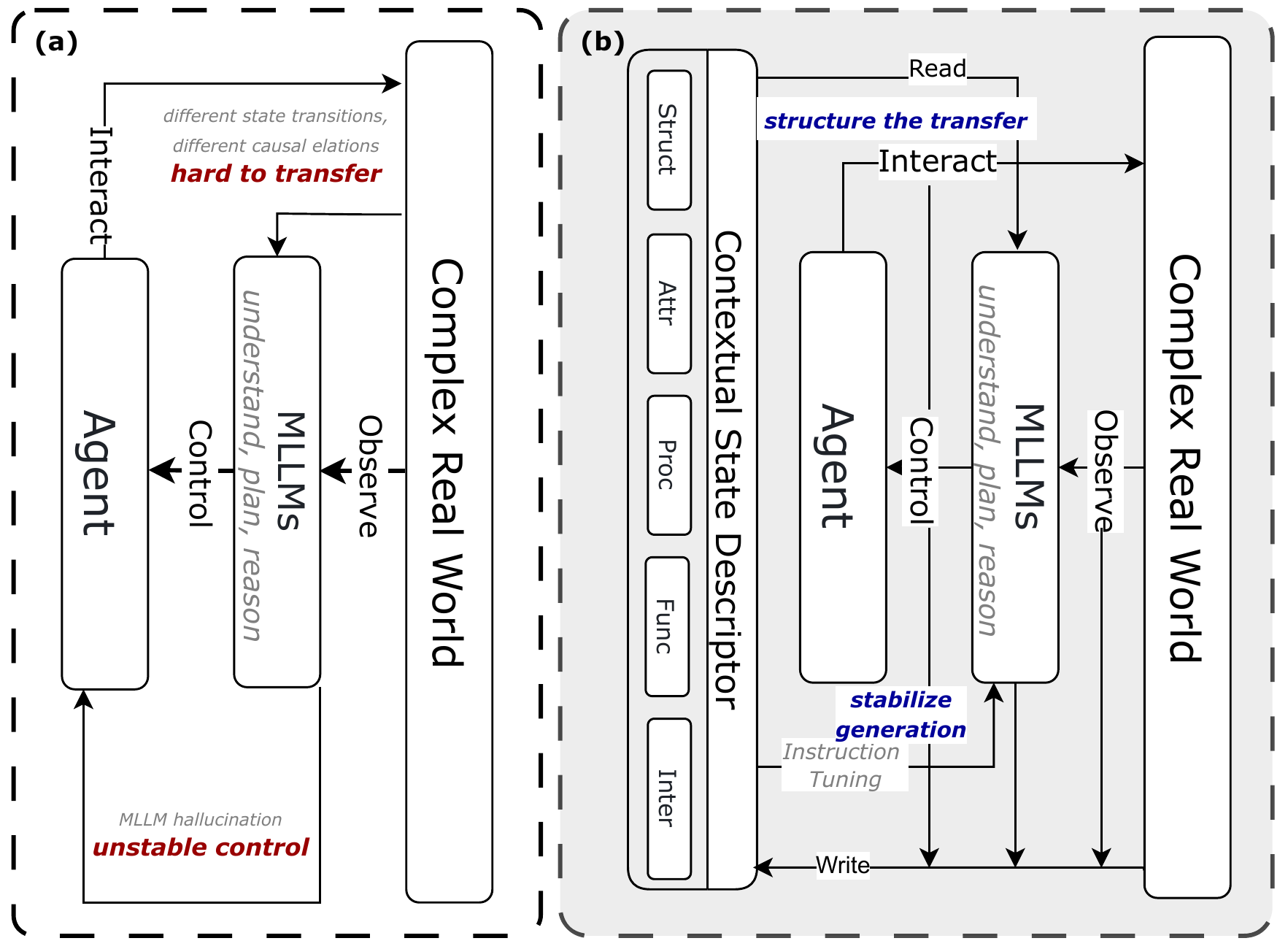}
    \caption{
    \textbf{Overview of motivation and Problem Framework.}
    (a) Traditional MLLM-based agents struggle to generalize across complex real-world environments due to different state transitions and causal relations (\textit{hard to transfer}) and may exhibit unstable control arising from hallucinations.
    (b) The proposed Structured In-Context Learning framework introduces a unified CSD that decomposes environmental knowledge into five explicit transfer dimensions.
    }
    \label{fig:motivation}
\end{figure}

As illustrated in Figure~\ref{fig:motivation}, these challenges can lead to unstable control and limited transferability in traditional MLLM-based agents when deployed in complex real-world environments. To address these issues, we propose a Structured ICL framework~\citep{hu2023context} based on explicit transfer dimensions (\S\ref{sec:etd}). This framework decomposes environmental representations and historical experiences into multiple dimensions within the multimodal space, explicitly modeling task transfer relationships along five axes: Structural, Attribute, Procedural, Functional, and Interaction.
By modeling these dimensions explicitly, the model can semantically interpret the correspondence and similarity between tasks, thereby enabling interpretable cross-task alignment and analogy-based reasoning.

The entire process is grounded in a unified CSD (\S\ref{sec:csd}) that integrates heterogeneous environmental and experiential information. Building upon this representation, an instruction-tuned MLLM performs structured reasoning and planning, achieving efficient and stable generalization to new tasks without parameter updates.

\subsection{Explicit Transfer Dimensions}
\label{sec:etd}

This section anchors the five ``learning-by-analogy'' dimensions within a unified, MLLM-driven framework for explicit knowledge transfer. In this formulation, a single MLLM serves as the central mechanism to represent, align, and evaluate cross-modal correspondences along five interpretable axes: \textit{structural, attribute, procedural, functional, and interaction}.

These five dimensions are not arbitrary-they form a holistic grammar of understanding for any agent operating in open-world environments. From a theoretical standpoint, an embodied agent seeking to transfer or reconstruct knowledge must simultaneously grasp three fundamental questions:

\begin{itemize}
    \item \textbf{What the world is like}---captured through its structures and attributes;
    \item \textbf{How the world changes}---revealed through its procedures and functions;
    \item \textbf{How the agent engages with the world}---embodied in its interactions.
\end{itemize}

These aspects are deeply interdependent and hierarchically organized. The structural and attribute dimensions ground understanding in the physical and spatial regularities of the environment, providing the static scaffolding upon which transfer can occur. The procedural and functional dimensions capture the world’s dynamics---its causes, effects, and transformations---enabling the reuse and adaptation of strategies across contexts. Finally, the interaction dimension closes the loop, modeling the continuous feedback between perception and action that allows an agent to learn not only from observation but from participation. Although our approach appears similar to MrSteve’s ``What-Where-When Memory,''~\citep{park2025mrsteve} MrSteve primarily aims to introduce episodic memory into agents, whereas our method focuses on leveraging structured memory to enable task transfer.

\begin{itemize}
\item \textit{Structural Axis---``How the world is organized.''}
Focuses on spatial layout and hierarchical relationships---how entities are arranged and connected.
It provides the geometric framework of the environment, helping the agent understand spatial structure, reachability, and coherence during transfer.

\item \textit{Attribute Axis---``What physical properties things have.''}
Describes the visual and physical traits of objects, such as color, texture, hardness, and material composition.
It enables reasoning about substitution, support, and compatibility in different contexts.

\item \textit{Procedural Axis---``How the world changes.''}
Captures the causal rules and state transitions that define how actions alter the environment.
It models sequences and dependencies---clarifying what to do, when, and why.

\item \textit{Functional Axis---``What things do.''}
Describes the purpose and role of objects---what they can do and how they contribute to tasks.
It supports semantic-level generalization and creative reuse across domains.

\item \textit{Interaction Axis---``How the agent interacts with the world.''}
Characterizes the perception–action loop, showing how operations lead to feedback and environmental response.
It connects knowledge to execution, ensuring actions are both understandable and performable.
\end{itemize}

To enable the agent to better transfer knowledge across the five dimensions, we design a standardized CSD-ICAL paradigm (\S~\ref{sec:csd}).

\subsection{Contextual State Descriptor}
\label{sec:csd}

This section introduces the overall design goals, data schema, generation pipeline, and system contracts of the CSD.
The core idea of CSD is to compress heterogeneous multimodal inputs---visual, textual, and interactive---into a unified, comparable, and verifiable semantic snapshot.
This unified representation enables Echo to perform stable retrieval and reasoning within a structured semantic space.
Specifically, the CSD organizes multimodal content along explicit transfer dimensions, including structural, attribute, procedural, functional, and interaction axes.
By aligning these axes through vectorized encoding and quality evaluation, the CSD serves as a reliable foundation for interpretable cross-task generalization and anomaly detection.
\begin{figure}[t!]
    \centering
    \includegraphics[width=\linewidth]{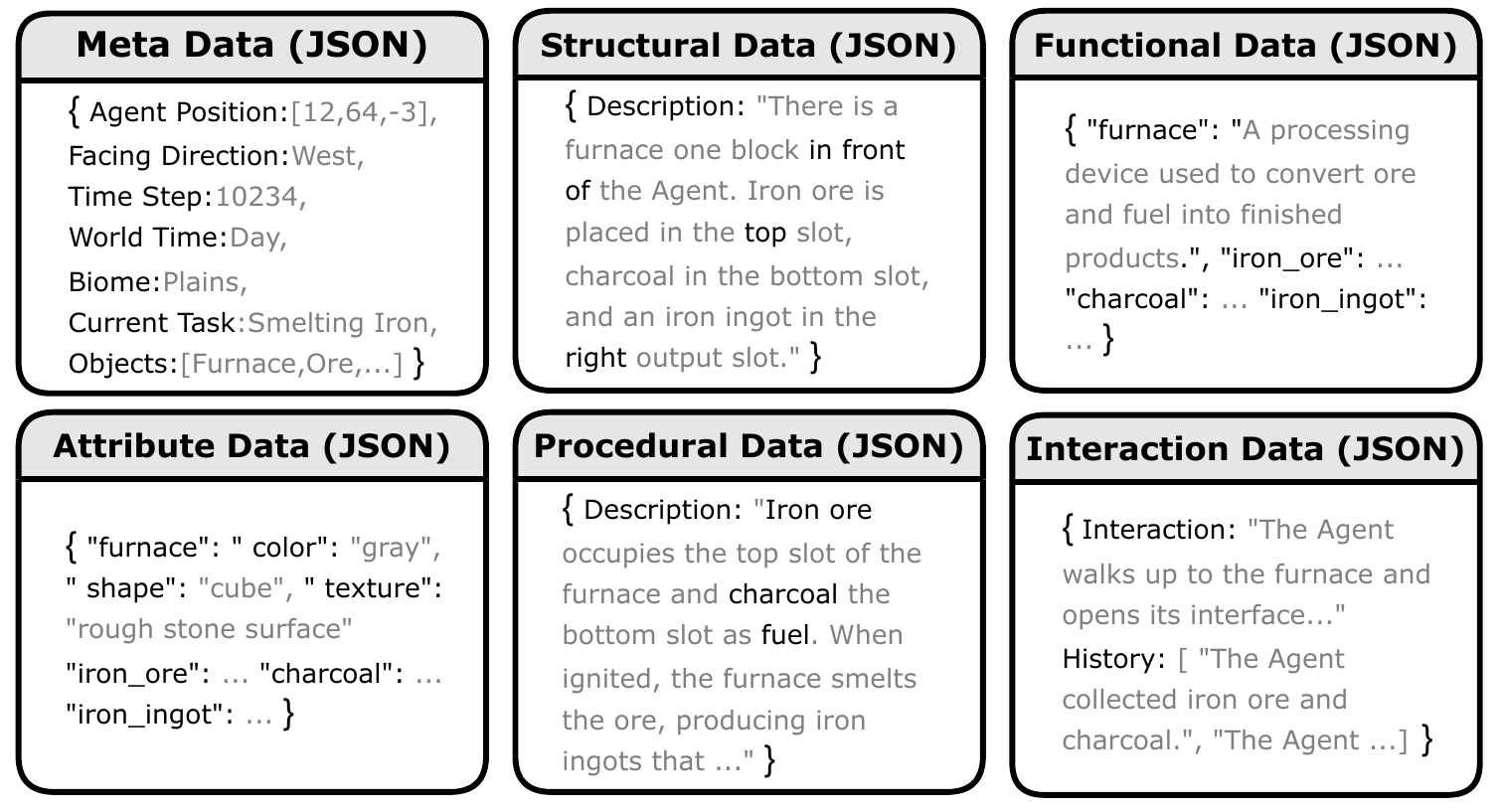}
    \caption{\textbf{Overview of the CSD schema.}}
    \label{fig:csd}
\end{figure}
\\ \noindent\textbf{Design Goals and Constraints.}
A CSD consists of six core components: metadata and five semantic dimensions.
The meta field records the generation timestamp, source environment, and model versions.
The five fields---struct, attr, proc, func, and inter---correspond to the five transfer dimensions, as illustrated in Fig.~\ref{fig:csd}, each containing symbolic content as well as global embeddings for fast vector-based retrieval.
In addition, during training, we apply instruction fine-tuning to enable the MLLM to produce well-formatted CSD structures more reliably.
Throughout this process, the model learns from large numbers of structured task examples to align task descriptions with evidence across the five semantic axes, generating normalized outputs that follow a unified specification.
The training data consists of multimodal task instructions, historical execution traces, and verifier feedback, enabling the model to develop basic capabilities in comprehension, structured organization, and consistency assurance when generating CSD.\\
\noindent\textbf{ICL-Based Analogical Learning.}
\begin{figure}[t!]
    \centering
    \includegraphics[width=0.9\linewidth]{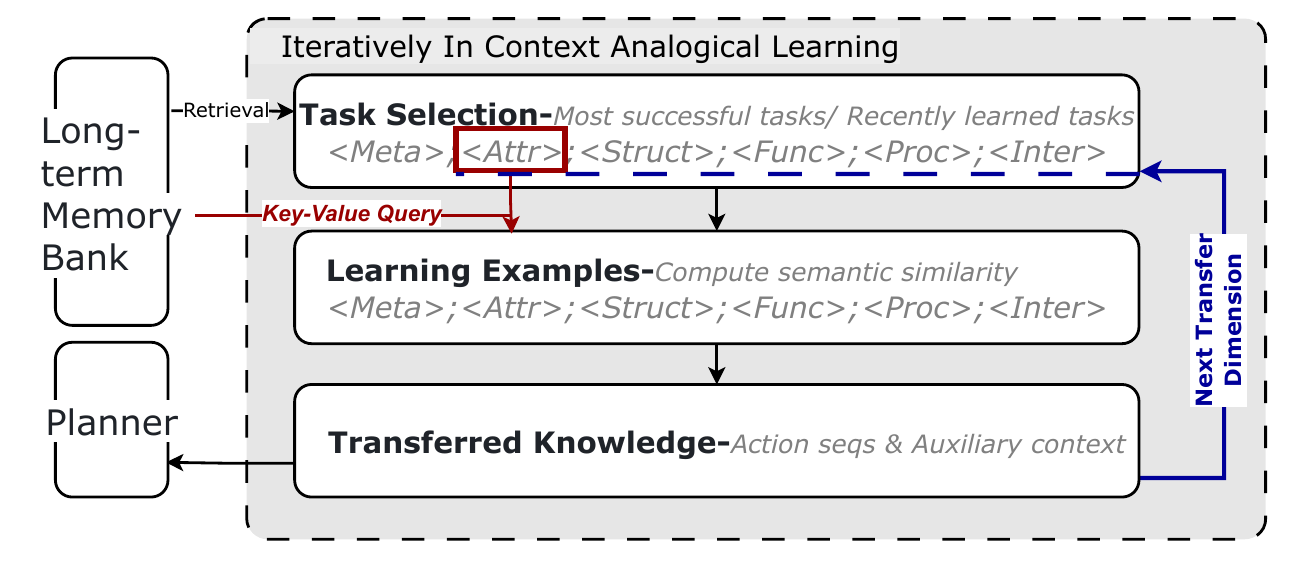}
    \caption{\textbf{ICL-based analogical learning workflow} using the CSD memory bank.}
    \label{fig:transfer}
\end{figure}
As illustrated in Fig.~\ref{fig:transfer},
CSD entries are aligned with the Minecraft task workflow:
only successful tasks are written to long-term memory.
The CSD library is periodically maintained offline through consolidation, cleaning,
deduplication, and clustering. Clustered CSD supports knowledge inference and pattern
abstraction (e.g., extending ``smelting iron ore $\rightarrow$ iron ingot''
to ``smelting gold ore $\rightarrow$ gold ingot'' or deriving new crafting routes),
enabling autonomous knowledge expansion.
Based on this representation, we build an ICL-based analogical workflow.
(1) \textbf{Task Selection}: choose a representative task
(e.g., most successful or most recently learned) and extract its complete CSD;
(2) \textbf{Example Retrieval}: retrieve the top-$K$ most relevant tasks
by computing multi-dimensional semantic similarity across five CSD components
(attr, struct, func, proc, Inter);
(3) \textbf{ICL Context Construction}: combine these samples to form
the ICL input context;
(4) \textbf{New Task Induction}: the model generalizes from the context
and outputs potential new tasks only as action sequences;
(5) \textbf{Execution \& Validation}: the actions are executed and evaluated;
successful trajectories are stored, while failures are logged.
This workflow enables continuous experience accumulation,
knowledge transfer, and autonomous task discovery.

\subsection{Overall Iterative Process}
\label{sec:oip}

\begin{figure}[t!]
    \centering
    \includegraphics[width=\linewidth]{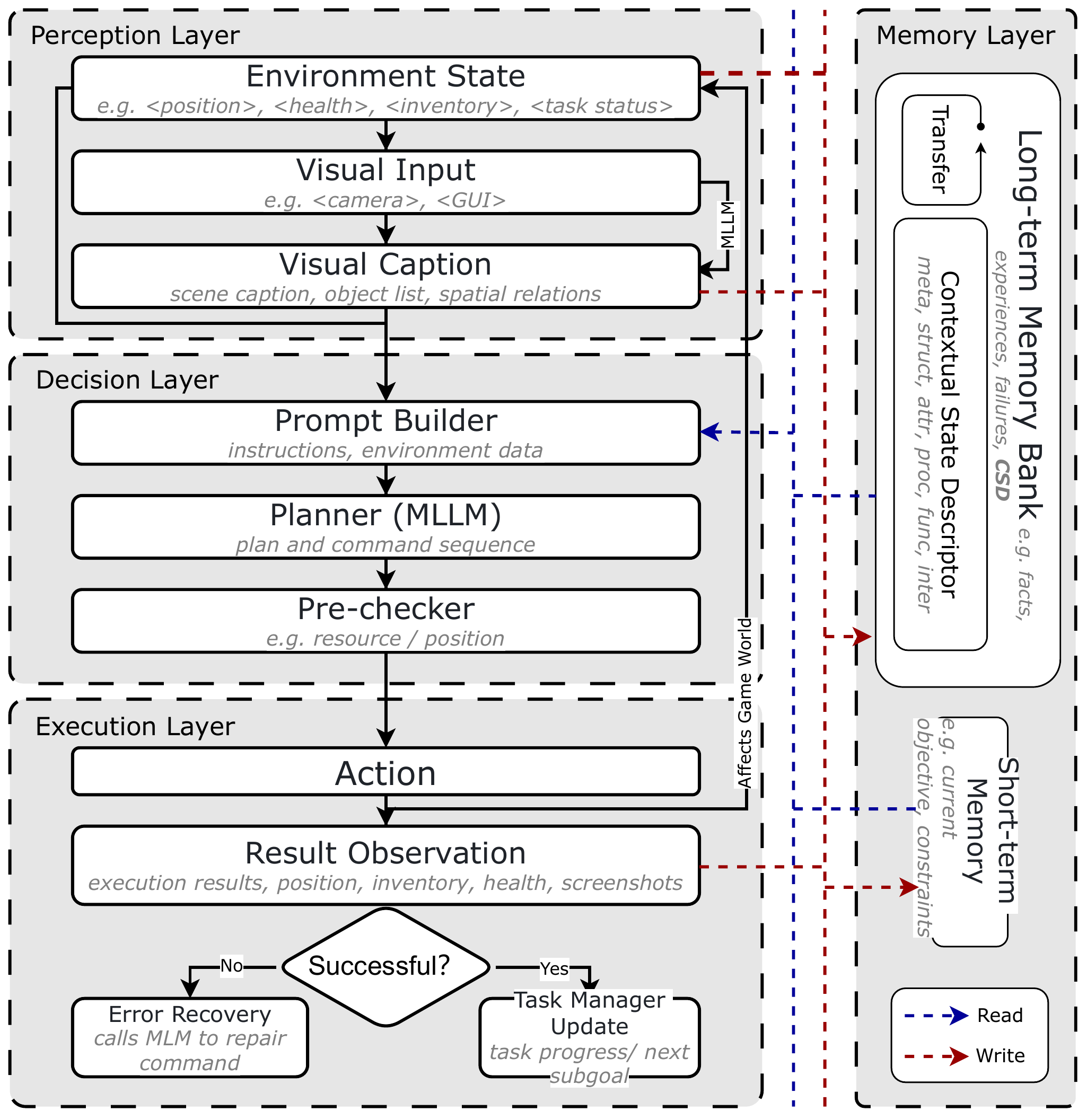}
    \caption{
    \textbf{Overview of our iterative framework.} The system performs
    perception, memory retrieval, planning, verification, and execution
    in a loop. A three-layer architecture (perception, decision, execution)
    interacts with short- and long-term memory to support structured ICAL
    and case-based transfer.
    }
    \label{fig:overall-architecture}
\end{figure}

\begin{table*}[h]
\centering
\small
\setlength{\tabcolsep}{5pt}
\renewcommand{\arraystretch}{1.1}
\caption{\textbf{From-scratch learning in Minecraft (Success@0→10 / Success@0→30)}. Higher is better. Results are averaged over worlds, map variants, and resource configurations. \faIcon{toggle-on} denotes full model; \faIcon{toggle-off} denotes component disabled.}
\resizebox{\linewidth}{!}{
\input{tab/main_tab}
}
\label{tab:cross_world_scratch}
\end{table*}

The system follows the classical agent model~\citep{wang2023voyager,wang2024jarvis} and achieves efficient use of internal knowledge for open-world task transfer by combining ICAL with explicit transfer axes, thereby improving the agent's performance and capabilities in the open world, as shown in Figure~\ref{fig:overall-architecture}.

\noindent\textbf{Transfer System Formalization.}
The iterative reasoning process is formalized as follows:
(1) \textbf{Memory:} The memory $\mathcal{M} = \{m_i\}$ stores multimodal trajectories, their CSDs, plans, validation results, and execution traces. It maintains dual representations: symbolic graphs for interpretability and vector embeddings for fast retrieval.
(2) \textbf{Transfer Space (Five Axes):} The transfer space $\mathbb{T} = \{\text{struct}, \text{attr}, \text{proc}, \text{func}, \text{inter}\}$ represents the structural, attribute, procedural, functional, and interaction dimensions of transferable knowledge.
(3) \textbf{Retrieval Operator:} The retrieval operator $\mathcal{S}_K = \mathcal{R}(x_t, \mathcal{M}, \mathbb{T})$ retrieves $K$ exemplars and their corresponding cross-axis similarity evidence from the memory bank.
(4) \textbf{Instruction-tuned MLLM:} The instruction-tuned MLLM, denoted as $f_\theta$ (with frozen parameters $\theta$), performs structured in-context learning:
\begin{equation}
    [\pi_t, \mathcal{A}\!s\!s_t] = f_\theta(x_t, \mathcal{S}_K, \text{protocol}),
    \label{eq:agent}
\end{equation}
where $\pi_t$ represents a hierarchical plan, and $\mathcal{A}\!s\!s_t$ corresponds to self-verification assertions.
(5) \textbf{Verifier:} The verifier $\{\text{pass}, \text{fail}\} = \mathcal{V}(\pi_t, \mathcal{A}\!s\!s_t, x_t), \text{ report}$ ensures the internal logical consistency and external task feasibility of the plan and assertions.
(6) \textbf{Executor:} The executor $\text{Exec}(\pi_t) \rightarrow \text{trace}_t$ executes the plan and collects the resulting trajectory.
(7) \textbf{Memory Update:} The memory update function $\mathcal{M}' = \mathcal{U}(\mathcal{M}, \text{trace}_t)$ updates both symbolic and vector channels for continual learning.

%% file: tab/main_tab.tex
\begin{tabular}{l|ccc|ccc|ccc|ccc}
\toprule
\multirow{2}{*}{\textbf{Method}} &
\multicolumn{3}{c}{\textbf{Recipe (Succ@0→10 / 30) $\uparrow$}} &
\multicolumn{3}{c}{\textbf{Functional Eq. (Succ@0→10 / 30) $\uparrow$}} &
\multicolumn{3}{c}{\textbf{Crafting Chain (Succ@0→10 / 30) $\uparrow$}} &
\multicolumn{3}{c}{\textbf{Utility Blocks (Succ@0→10 / 30) $\uparrow$}} \\
\cmidrule{2-13}
 & Bed & Iron Pickaxe & Shield &
   BridgeEq & SmeltEq & WeaponEq &
   WeaponSet & ToolBench & ArmorSet &
   CraftGrid & CraftTable & Furnace \\
\midrule
\multicolumn{13}{c}{\textbf{Memory-enhanced Models}} \\
\midrule
\faIcon{toggle-on} Voyager~\citep{wang2023voyager} & 35.0 / 62.5 & 30.0 / 57.5 & 25.0 / 50.0 & 17.5 / 40.0 & 15.0 / 35.0 & 15.0 / 32.5 & 22.5 / 45.0 & 27.5 / 55.0 & 17.5 / 40.0 & 45.0 / 77.5 & 35.0 / 65.0 & 25.0 / 47.5 \\
\faIcon{toggle-off} \textit{Self Verification}~\citep{wang2023voyager} & 25.0 / 47.5 & 25.0 / 45.0 & 17.5 / 37.5 & 12.5 / 27.5 & 10.0 / 25.0 & 7.5 / 20.0 & 15.0 / 32.5 & 17.5 / 40.0 & 10.0 / 27.5 & 30.0 / 55.0 & 22.5 / 45.0 & 12.5 / 32.5 \\
\faIcon{toggle-on} MrSteve~\citep{park2025mrsteve} & 22.5 / 40.0 & 20.0 / 37.5 & 17.5 / 35.0 & 47.5 / 77.5 & 45.0 / 77.5 & 42.5 / 72.5 & 15.0 / 32.5 & 17.5 / 35.0 & 12.5 / 27.5 & 25.0 / 45.0 & 17.5 / 35.0 & 10.0 / 22.5 \\
\faIcon{toggle-on} MP5~\citep{qin2024mp5} & 40.0 / 67.5 & 37.5 / 65.0 & 35.0 / 60.0 & 45.0 / 75.0 & 42.5 / 70.0 & 37.5 / 65.0 & 35.0 / 65.0 & 35.0 / 62.5 & \textbf{30.0} / 57.5 & 42.5 / 72.5 & 35.0 / 65.0 & 27.5 / 52.5 \\
\faIcon{toggle-off} \textit{Patroller}~\citep{qin2024mp5} & 35.0 / 55.0 & 32.5 / 52.5 & 27.5 / 47.5 & 35.0 / 60.0 & 35.0 / 55.0 & 30.0 / 50.0 & 27.5 / 50.0 & 25.0 / 45.0 & 22.5 / 42.5 & 35.0 / 57.5 & 30.0 / 50.0 & 22.5 / 37.5 \\
\faIcon{toggle-on} JARVIS-1~\citep{wang2024jarvis} & 60.0 / 87.5 & 50.0 / 85.0 & 50.0 / 80.0 & 47.5 / 77.5 & 45.0 / 75.0 & 40.0 / 70.0 & 37.5 / 72.5 & \textbf{42.5} / 75.0 & \textbf{30.0} / 65.0 & 52.5 / 80.0 & \textbf{55.0} / 82.5 & 35.0 / 67.5 \\
\faIcon{toggle-off} \textit{SelfCheck}~\citep{wang2024jarvis} & 45.0 / 75.0 & 42.5 / 70.0 & 37.5 / 65.0 & 35.0 / 60.0 & 32.5 / 57.5 & 27.5 / 50.0 & 25.0 / 50.0 & 27.5 / 57.5 & 17.5 / 40.0 & 45.0 / 72.5 & 32.5 / 60.0 & 20.0 / 45.0 \\
\midrule
\multicolumn{13}{c}{\textbf{Few-shot Variants}} \\
\midrule
\faIcon{toggle-on} Echo (1-shot) & 50.0 / 80.0 & 50.0 / 80.0 & 42.5 / 75.0 & 37.5 / 60.0 & 32.5 / 65.0 & 27.5 / 50.0 & 37.5 / 72.5 & 32.5 / 67.5 & 20.0 / 60.0 & 55.0 / \textbf{92.5} & \textbf{55.0} / 75.0 & 35.0 / 62.5 \\
\faIcon{toggle-on} Echo (2-shot) & \textbf{62.5} / 90.0 & 50.0 / \textbf{87.5} & 52.5 / 80.0 & 35.0 / 65.0 & 37.5 / 62.5 & 40.0 / 65.0 & 37.5 / 72.5 & 40.0 / 70.0 & 22.5 / \textbf{67.5} & 55.0 / \textbf{92.5} & \textbf{55.0} / \textbf{87.5} & 35.0 / 62.5 \\
\faIcon{toggle-on} Echo (4-shot) & \textbf{62.5} / \textbf{92.5} & 50.0 / \textbf{87.5} & 52.5 / 85.0 & 40.0 / 72.5 & 35.0 / 62.5 & 40.0 / 65.0 & 37.5 / 72.5 & 40.0 / 75.0 & 22.5 / \textbf{67.5} & 55.0 / \textbf{92.5} & \textbf{55.0} / \textbf{87.5} & 35.0 / 65.0 \\
\faIcon{toggle-on} Echo (8-shot) & \textbf{62.5} / \textbf{92.5} & \textbf{52.5} / \textbf{87.5} & \textbf{55.0} / \textbf{87.5} & \textbf{50.0} / \textbf{80.0} & \textbf{47.5} / \textbf{80.0} & \textbf{45.0} / \textbf{75.0} & \textbf{40.0} / \textbf{77.5} & \textbf{42.5} / \textbf{82.5} & 27.5 / \textbf{67.5} & \textbf{57.5} / \textbf{92.5} & \textbf{55.0} / \textbf{87.5} & \textbf{37.5} / \textbf{70.0} \\
\bottomrule
\end{tabular}

%% file: sec/4_experiment.tex
\section{Experiment}
\label{sec:exp}

\noindent\textbf{Experimental Objectives.} Integrate multimodal long-term memory with transfer learning to address the open-world distribution shift problem. (1) To demonstrate that explicit transfer axes combining structured ICAL outperform methods based solely on memory or policy transfer in cross-world and cross-task generalization;
(2) To show that consistency self-checking contributes to stability in long-horizon planning;
(3) To illustrate how memory replay (continual learning) leads to progressive performance improvement and enhanced interpretability.

\subsection{Cross-world Learning from Scratch}
\label{sec:mainexp}

\noindent\textbf{Multi-task Evaluation Under Cold-starting.} As shown in Table~\ref{tab:cross_world_scratch}, the primary objective of this section is to evaluate an agent’s generalization ability and learning efficiency from scratch under open-world testing. Specifically, this experiment examines whether a model, when initialized with no prior task knowledge, can rapidly acquire crafting and reasoning skills within the first 10 and 30 learning episodes.
\\ \noindent\textbf{Evaluation Tasks and Metrics.} The tasks are grouped into four families:
Recipe (e.g., Iron Pickaxe, Bed, Shield) evaluates structural and shape-based recipe transfer.
Functional Equivalence (BridgeEq, SmeltEq, WeaponEq) tests the agent’s ability to perform functionally equivalent reasoning---i.e., using alternative equivalent items when the required ones are unavailable.
Crafting Chain (WeaponSet---crafting wooden, stone, and iron swords; ToolBench---crafting a stone pickaxe, shovel, and hoe; ArmorSet---crafting a full set of iron armor) measures multi-step dependency reasoning.
Finally, Utility Blocks (Crafting Grid, Crafting Table, Furnace) evaluate the agent’s ability to correctly utilize functional blocks to complete short-horizon, dependency-based tasks.
Success@0$\rightarrow$10 and Success@0$\rightarrow$30 represent average success rates within the first 10 and 30 episodes, respectively (higher is better).
\\ \noindent\textbf{2-Shot is Competitive.} The comparison includes several strong baselines and ablation variants: Voyager~\citep{wang2023voyager}, MrSteve~\citep{park2025mrsteve}, MP5~\citep{qin2024mp5}, and JARVIS-1~\citep{wang2024jarvis}, each with corresponding component removal experiments. Voyager~\citep{wang2023voyager} without the Self-Verification mechanism shows a significant drop in performance, indicating that self-verification is crucial for stable learning. MrSteve performs best on functional-equivalence tasks but is weaker on structural and multi-step reasoning tasks. MP5 demonstrates consistent robustness, and its Patroller module (inspection module)~\citep{qin2024mp5} significantly improves task success rates under complex visual variations. JARVIS-1 stands out as the most stable overall baseline, achieving leading scores across nearly all task families. Removing its SelfCheck module~\citep{wang2024jarvis} leads to a 10--20 point drop in success rates for most tasks, further confirming that self-consistency checking is vital for cross-world stability. Echo exhibits competitive performance even in the 2-shot setting.

\noindent\textbf{Few-Shot Gains Saturate.} In the few-shot setting, our method introduces ICAL during planning with 1--8 contextual examples. Results show steady improvement as the number of examples ($k$) increases, especially for multi-step and long-term tasks. The 1-shot variant already matches or surpasses most baselines, while the 4-shot and 8-shot variants achieve the highest overall success rates---up to 62.5/92.5 on the Recipe and Crafting Table families. Although functional-equivalence tasks in the 8-shot setting do not reach JARVIS-1~\citep{wang2024jarvis}’s absolute peak, their learning curves are smoother, indicating that ICAL effectively accelerates cross-world adaptation during cold-start learning.

\begin{figure*}
    \centering
    \includegraphics[width=\linewidth]{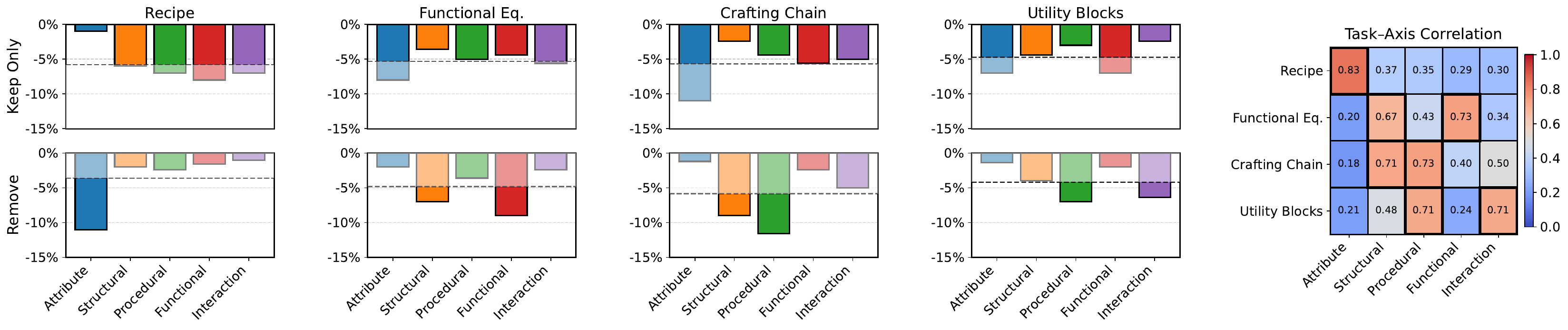}
    \caption{
        \textbf{Comparison of task performance when keeping or removing individual design axes.}
        Left: bar charts for each task showing median performance change under
        ``Keep Only'' and ``Remove'' scenarios. Right: correlation heatmap between
        task outcomes and design axes (thicker borders indicate stronger correlations).
    }
    \label{fig:ablation}
\end{figure*}

\subsection{Continuous Learning Test}

\begin{figure}[t!]
    \centering
    \includegraphics[width=\linewidth]{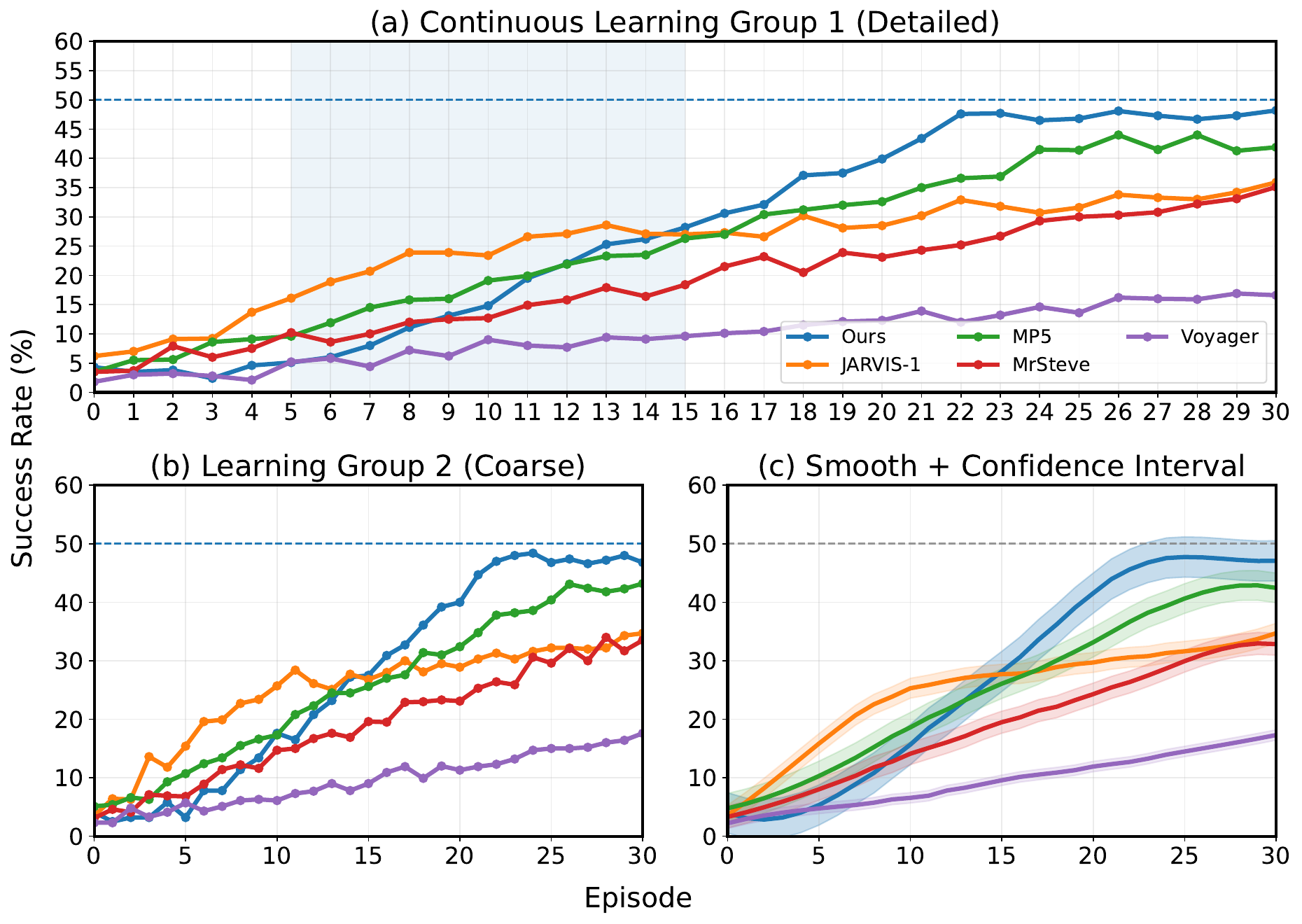}
    \caption{
    \textbf{Continuous learning performance comparison.}
    The figure shows the success rate (\%) over 31 training episodes (0--30) across five agents: \textbf{Ours}, \textbf{JARVIS-1}, \textbf{MP5}, \textbf{MrSteve}, and \textbf{Voyager}.
    The shaded region (episodes 5--15) highlights the fast learning phase of our method.
    Compared to all baselines, \textbf{our method demonstrates a faster learning rate in the mid-phase}.
    }
    \label{fig:learning_curve}
\end{figure}

This experiment (Fig.~\ref{fig:learning_curve}) aims to evaluate the proposed method's continuous learning efficiency and stability under open-world tasks. Specifically, we focus on the following aspects: (1) the adaptation speed during the early cold-start phase (episodes 0--10); (2) the improvement and stability during the mid-to-late phase (episodes 20--30) driven by long-horizon planning and memory replay; and (3) the crossover points and final convergence performance compared with strong baselines, including JARVIS-1, MP5, MrSteve, and Voyager. The goal is to validate the effectiveness of our method in achieving sustainable adaptation and cross-task generalization in long-term learning scenarios.

\noindent\textbf{Mid-to-Late Phase Performance.}
Our method shows slower progress in the early episodes but accelerates rapidly after episode 10, reaching a stable plateau around 46--48\% in the final stage. In contrast, JARVIS-1 learns faster initially but saturates after 20 episodes, while MP5 improves steadily, and both MrSteve and Voyager remain at lower performance levels. At episode 30, the success rate ranking is: Echo (45), MP5 (43)~\citep{qin2024mp5}, JARVIS-1 (35)~\citep{wang2024jarvis}, MrSteve (33)~\citep{park2025mrsteve}, and Voyager (18)~\citep{wang2023voyager}.
From a ``fast start vs. strong finish'' perspective, JARVIS-1~\citep{wang2024jarvis} excels in the cold-start phase (0--10) due to its pre-trained policy library and self-checking mechanism, but its growth plateaus beyond 20 episodes, indicating limited scalability in multi-task and long-horizon reasoning.
Although ours starts slower, it maintains steady gains from episodes 10--30, validating that explicit transfer axes (Attr/Struct/Proc/Func/Inter) and structured ICL foster stronger long-term planning and knowledge reuse once sufficient experience is accumulated. This trend is consistent with the task-family sensitivity and stability analyses in \S\ref{sec:mainexp} and \S\ref{sec:ablation}.

\subsection{Ablation of Explicit Transfer Axes}
\label{sec:ablation}

\noindent\textbf{Single-Axis Removal/Retention.}
As shown in Figure~\ref{fig:ablation}, this experiment aims to verify the necessity and advantages of using explicit transfer axes in multi-axis similarity modeling. By comparing two ablation settings-Keep-Only (retaining only one axis) and Remove (removing one axis)---we quantify the independent contribution and sensitivity of each semantic axis across four task families (Recipe, Functional Eq., Crafting Chain, Utility Blocks). The goal is to examine whether explicit axis alignment provides higher stability and interpretability than implicit holistic similarity modeling.
The setup involves four task families and five explicit transfer axes, resulting in ten ablation scenarios: five Remove-Axis (removing one axis) and five Keep-Only-Axis (retaining only one axis). The evaluation metric is the relative change in success rate ($\Delta$Success\%), where negative values indicate performance degradation.
\\ \noindent\textbf{Axis Interactions.} The experimental results show that different semantic transfer axes have a significant impact on task performance:
\begin{itemize}
    \item The \textit{Attribute axis} is crucial for recipe tasks. Removing it leads to a significant drop in Recipe (-11\%).
    \item The \textit{Structural axis} affects functional equivalence and crafting chain tasks. Removing it causes declines in Functional Eq. (-7\%) and Crafting Chain (-9\%).
    \item The \textit{Procedural axis} has the greatest impact on long-horizon tasks (e.g., Crafting Chain). Removing it leads to severe degradation (-12\%).
    \item The \textit{Functional axis} dominates functional equivalence tasks. Removing it almost disables the tasks (-9\%).
    \item The \textit{Interaction axis} affects short-term tasks. Removing it causes a major drop in Utility Blocks (-7\%).
\end{itemize}
The experimental results show that removing a single axis causes a significant performance drop, much greater than the drop when only retaining a single axis. The Structural axis primarily affects geometric and recipe-related tasks; the Attribute axis is crucial for functional equivalence and crafting chain tasks; the Procedural axis has a decisive impact on long-horizon tasks like Crafting Chain and Crafting Table; the Functional axis dominates functional equivalence tasks; and the Interaction axis strongly influences utilities operation tasks. Explicit transfer axes help clarify the dependencies between tasks and effectively locate the sources of errors, such as process reasoning errors or interface positioning errors.

\subsection{Case Study}

This section demonstrates an experimental example for interpretability analysis. The target sample is retrieved and matched through the Func axe (since the functions of planks and stone are similar). After performing ICL, a similar example for the stone pickaxe is inferred. The specific process is shown in the Figure~\ref{fig:example}. The key CSD for the entire transfer process is shown below.

\noindent\textbf{Transfer Example.} Crafting a Wooden Pickaxe.

\noindent\textbf{Task Steps.}
(1) Convert oak logs into oak planks.
(2) Turn planks into sticks.
(3) Attempt to craft the pickaxe directly but fail, realizing that a crafting table is required.
(4) Craft and place the crafting table.
(5) Arrange planks and sticks on the crafting table to craft the wooden pickaxe.

\noindent\textbf{Retrieved Item Descriptions (Func Content).}
Oak Planks: Used as materials for crafting tools, and as building materials.
Stone: Serve as components for crafting tools and construction materials.

\noindent\textbf{ICAL Result.} Crafting a Stone Pickaxe.

\noindent\textbf{Task Steps.}
(1) Use the wooden pickaxe to mine stone blocks and acquire stone.
(2) Collect planks and craft planks into sticks.
(3) Craft and place the crafting table.
(4) Arrange stones and sticks on the crafting table to craft the stone pickaxe.

\noindent\textbf{Analysis.} The key point of this transfer is the functional similarity between materials (oak planks and stone), which allows the task structure for crafting a wooden pickaxe to be applied to crafting a stone pickaxe. ICAL recognizes the pattern in task steps (gather materials, use the crafting table, arrange materials) and adapts it, transferring the knowledge from one task to the other.

\begin{figure}[t!]
    \centering
    \includegraphics[width=\linewidth]{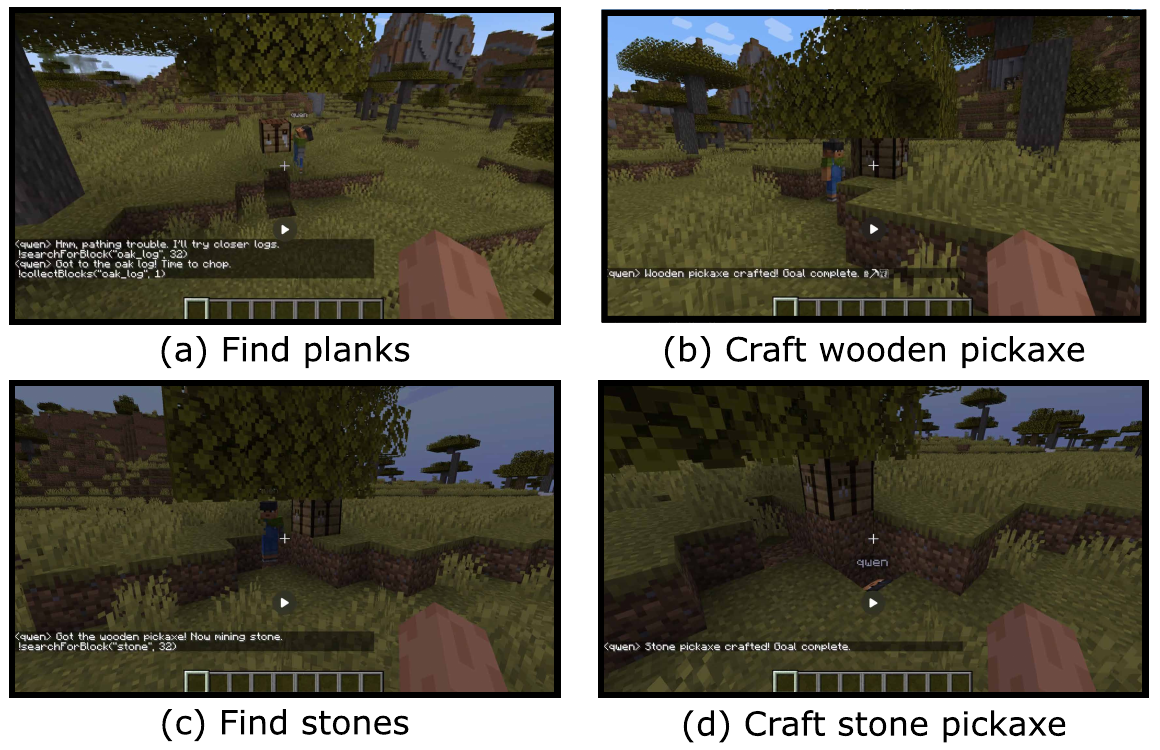}
    \caption{Transferring from a wooden pickaxe to a stone pickaxe.}
    \label{fig:example}
\end{figure}

%% file: sec/5_conclusion.tex
\section{Conclusion and Discussion}
\label{sec:con}

\noindent\textbf{Research Focus Comparison.}
Compared with representative works~\citep{wang2023voyager,wang2024jarvis,qin2024mp5}, our approach emphasizes skill acquisition and learning rather than exploration or perception. As a result, Echo is less effective in actively exploring unfamiliar environments. For instance, MP5 employs active perception to continuously gather new information~\citep{qin2024mp5}, whereas Echo relies more on prior knowledge and retrieval, which weakens its performance in information-sparse settings. In addition, our method shows a slower initial learning rate.
\\\noindent\textbf{Applicability to the Real Physical World.}
Our method is evaluated mainly in Minecraft, an open-ended and complex but still idealized environment with simple and consistent rules~\citep{mon2025embodied}. While such predictability facilitates efficient skill learning and transfer, it also limits real-world applicability. Compared with Minecraft, real-world tasks are more diverse, ambiguous, and causally complex, making transfer learning more reliant on the reasoning and generalization abilities of large models. Therefore, skill transfer in the real world is unlikely to be as straightforward as in Minecraft.
\\\noindent\textbf{Conclusion.}
This study explores explicit multimodal memory transfer in multimodal embodied agents. To address challenges in structuring and transferring multimodal memory, we propose five explicit transfer dimensions and integrate them with ICAL for effective organization and use. Compared with four baseline agents focusing on different aspects, our model achieves superior learning efficiency and task generalization in learning-from-scratch settings. We hope this framework inspires future research on enhancing planning and reasoning in interactive multimodal agents.
\\\noindent\textbf{Acknowledgments.}
This work was partially supported by the National Natural Science Foundation of China under grant 62572104, and 62220106008.